%% file: bmvc_final.tex
\documentclass[letterpaper]{bmvc2k}

\usepackage{tikz}
\usepackage{comment}
\usepackage{amsmath,amssymb} % define this before the line numbering.
\usepackage{color}
\usepackage{booktabs}
\usepackage{adjustbox}
\usepackage{bm}
\usepackage{multirow}
\usepackage{xcolor}
\usepackage{wrapfig}
\usepackage{url}
\title{Multi-View Neural Surface Reconstruction with Structured Light}

\addauthor{Chunyu Li}{chunyuli@preferred.jp}{1}
\addauthor{Taisuke Hashimoto}{hashimotot@preferred.jp}{1}
\addauthor{Eiichi Matsumoto}{matsumoto@preferred.jp}{1}
\addauthor{Hiroharu Kato}{hkato@preferred.jp}{1}

\addinstitution{
Preferred Networks, Inc.\\
3F Otemachi-building,\\ 1-6-1 Otemachi, Chiyoda-Ku,\\
Tokyo, Japan
}

\runninghead{C. Li ET AL}{Multi-View Neural Surface Reconstruction}

\begin{document}

\maketitle

\input{s0_abstract}
\input{s1_introduction}
\input{s3_method}
\input{s4_experiments}
\input{s5_conclusion}

\bibliography{egbib}
\end{document}

\maketitle
\setlength{\abovedisplayskip}{3pt}
\setlength{\belowdisplayskip}{3pt}

\section{Details on noise reduction}

\setlength\intextsep{0pt}
\begin{wrapfigure}[14]{R}[0pt]{0pt}
\centering
\includegraphics[width=0.45\textwidth]{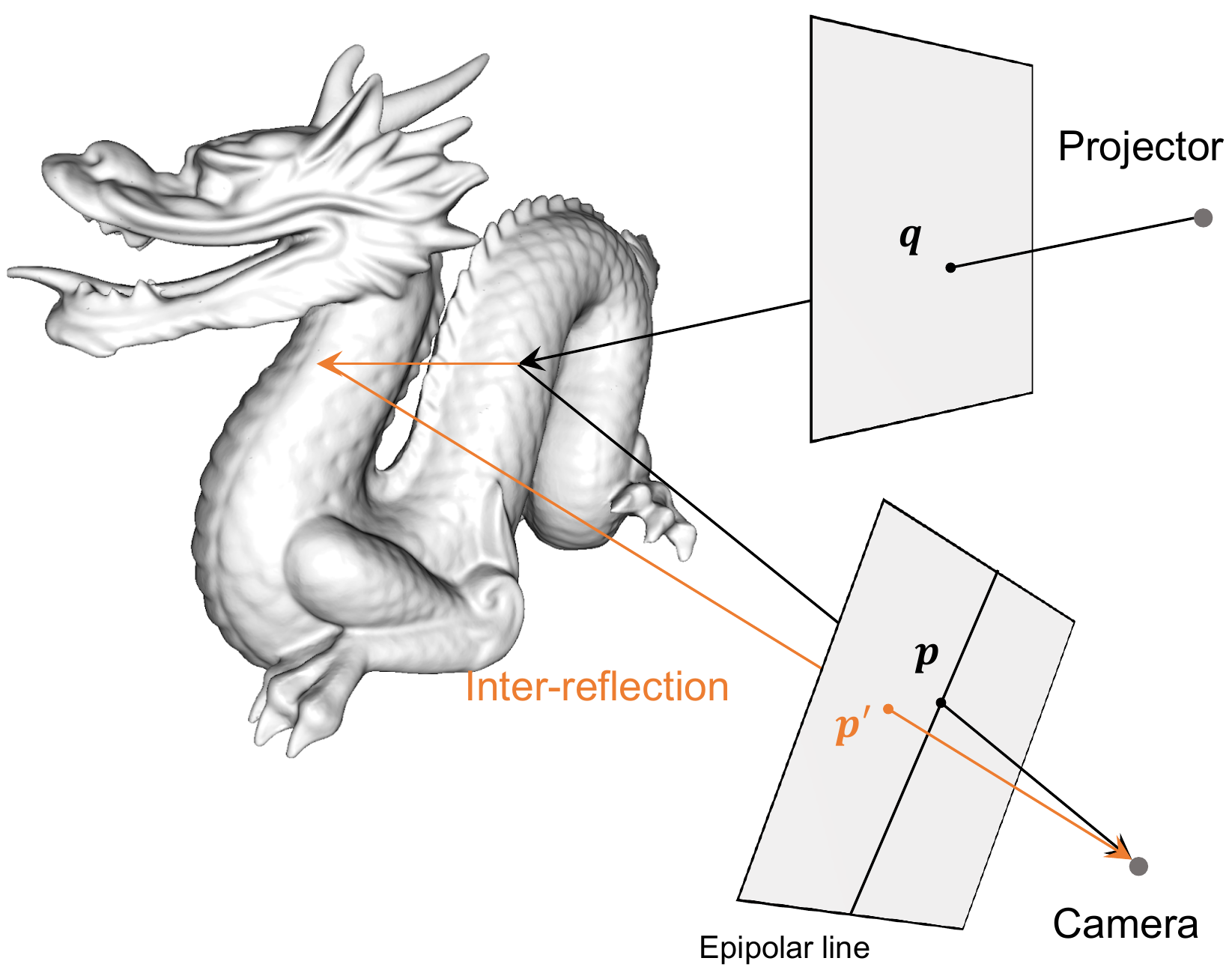}
\caption{Illustration of pattern misdetection caused by inter-reflection.}
\label{fig:noisereduction}
\end{wrapfigure}

As described in Section~2.1 of the main paper, we reduce the misdetection of the structured-light pattern caused by inter-reflection by calculating the epipolar line between the projecter and camera pair.
To be specific, as shown in Fig.~\ref{fig:noisereduction}, the light projected from the projector pixel $\bm{q}$ can reach the camera in one of two general ways: (1) by direct surface reflection, captured by a camera pixel $\bm{p}$ on the epipolar line (black path), which is the desirable path of the light for pattern decoding, or (2) by inter-reflection, captured by a camera pixel $\bm{p}^{\prime}$ that is not on the epipolar line (orange path).
Therefore, we can determine whether a decoded pixel is affected by inter-reflection using the epipolar line.
As the camera poses are unknown in our experiment, we calculate a rough fundamental matrix between the camera and projector from the noisy corresponding points using Ransac algorithm, and estimate the epipolar lines using this fundamental matrix.
Then, we eliminate correspondences whose camera pixels are not on the epipolar line.
Note that although we can effectively reduce most noise using this strategy, some limitations remain: (1) the estimated epipolar lines may include minor errors owing to the noisy corresponding points, and (2) we cannot eliminate the inter-reflected correspondences whose projector and camera pixels are on corresponding epipolar lines.
However, the amount of noise caused by these cases is small, so they can be further reduced by the photometric supervision introduced in Section~2.4 of the main paper.
The effectiveness of this noise-reduction strategy is demonstrated by the ablation study (see Section~\ref{ssec:ablationstudies} in supplementary material).

\section{Details on triangulation}
In this section we will explain the details on the calculation of $\bm{y}_a$ and $\bm{y}_b$ in Eq.~(5) of the main paper.
$\bm{y}_a$ and $\bm{y}_b$ are the nearest points between the two skew camera rays $R_a(\tau)$ and $R_b(\tau)$ (see the right column of Fig.~4).
We denote $R_a(\tau) = \left\{ \bm{o}_a + t_a\bm{v}_a \mid t_a \geq 0 \right\}$ and $R_b(\tau) = \left\{ \bm{o}_b + t_b\bm{v}_b \mid t_b \geq 0 \right\}$.
The cross product of $\bm{v}_a$ and $\bm{v}_b$ is perpendicular to the lines:
\begin{equation}
\bm{n} = \bm{v}_a \times \bm{v}_b.
\label{eq:product}
\end{equation}
The plane formed by the translations of $R_b(\tau)$ along $\bm{n}$ contains the point $\bm{o}_b$ and is perpendicular to $\bm{n}_1=\bm{v}_b \times \bm{n}$.
Therefore, the intersecting point of $R_a(\tau)$ with the above-mentioned plane, which is also the point on $R_b(\tau)$ that is nearest to $R_a(\tau)$, is given by
\begin{equation}
\bm{y}_a = \bm{o}_a + \frac{(\bm{o}_b-\bm{o}_a) \cdot \bm{n}_1}{\bm{v}_a \cdot \bm{n}_1}\bm{v}_a.
\label{eq:y_a}
\end{equation}
Similarly, the point on $R_b(\tau)$ nearest to $R_a(\tau)$ is given by
\begin{equation}
\bm{y}_b = \bm{o}_b + \frac{(\bm{o}_a-\bm{o}_b) \cdot \bm{n}_2}{\bm{v}_b \cdot \bm{n}_2}\bm{v}_b,
\label{eq:y_b}
\end{equation}
where $\bm{n}_2=\bm{v}_a \times \bm{n}$.

\section{Initial camera poses estimation for real-world dataset}
In the experiment on real-world scenes, the initial camera poses were measured using 26 AprilTag 16h5 Markers~\cite{richardson2013iros} fixed on the turntable.
We assume the intrinsic parameters of the cameras are known.
After capturing the multi-view input images, the initial camera poses are estimated following four steps.

\vspace{3mm}
\noindent \textbf{Step 1. Marker Detection:}
Given each image containing AprilTag 16h5 Markers, the detection process has to return a list of detected markers. Each detected marker includes the position of its four corners in the image and the id of the marker. This step is implemented using OpenCV ArUco module~\cite{aruco}.

\vspace{3mm}
\noindent \textbf{Step 2. Camera Pose Initialization:}
The next thing is to obtain the camera pose from detected markers.
First, for each image, the pose of each marker in the camera coordinate system is estimated individually using OpenCV ArUco module~\cite{aruco}.
Then using one marker as a reference , all camera poses in one coordinate system can be obtained by calculating the 3D transformation from each camera coordinate systems to the reference marker coordinate system.

\vspace{3mm}
\noindent \textbf{Step 3. Camera Pose Optimization:}
The camera poses obtained by Step 2 usually have large error. Next they are optimized using bundle adjustment while simultaneously updating the marker poses.
Specifically, our bundle adjustment jointly refining the camera poses and marker poses by minimizing the reprojection error of four corners of each marker.

\begin{figure}[t]
  \centering
\begin{minipage}[b]{0.455\hsize}\vspace{0pt}
  \centering
 \includegraphics[width=\hsize]{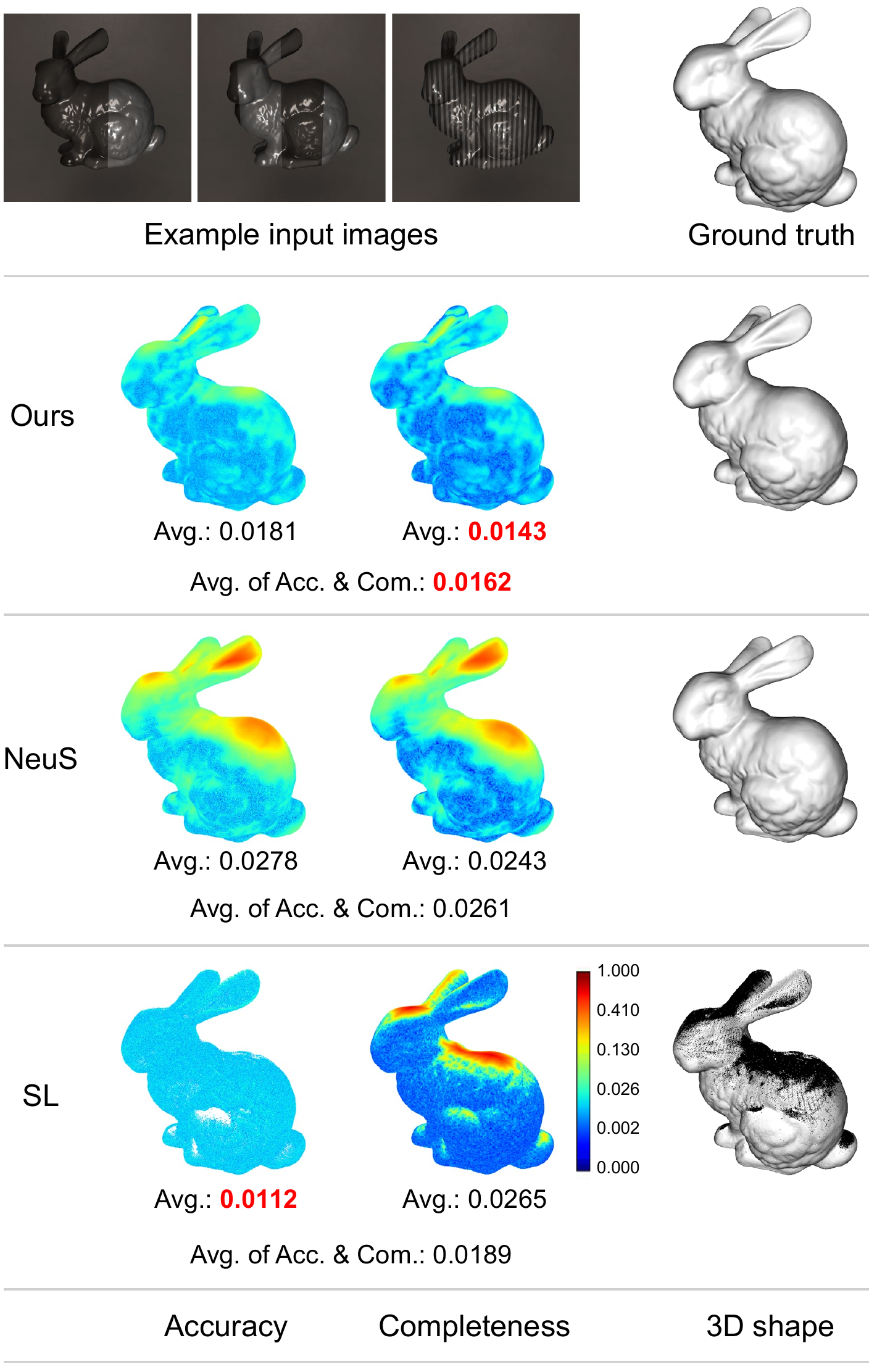}
 {\footnotesize (a) Stanford Bunny (glass)}\\
\end{minipage}
\hspace{10pt}
\begin{minipage}[b]{0.48\hsize}\vspace{0pt}
  \centering
 \includegraphics[width=\hsize]{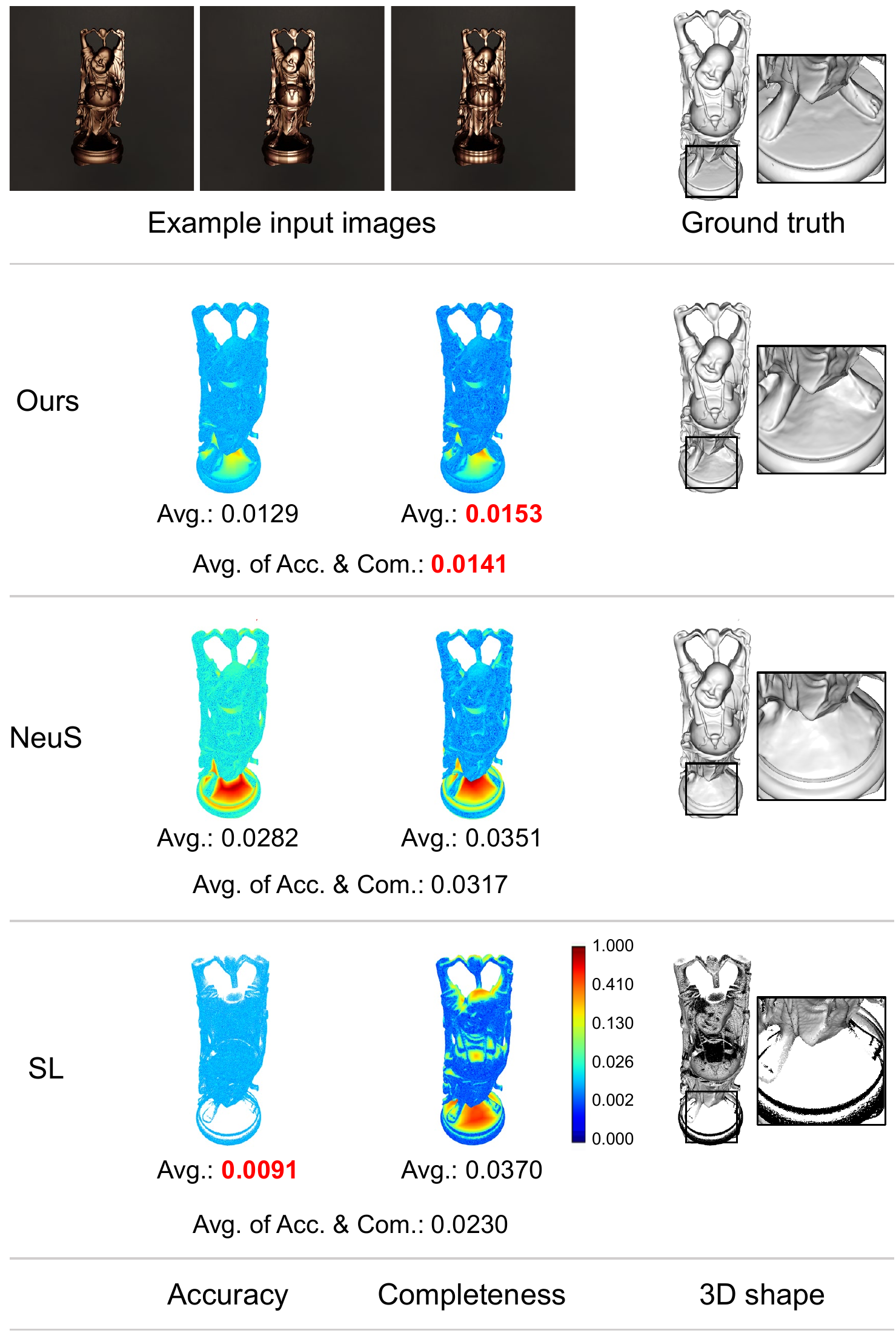}
 {\footnotesize (b) Happy Buddha (metal)}\\
\end{minipage}
 \caption{Example input images, 3D reconstruction results, and their completeness and accuracy errors on two additional synthetic scenes with {\it fixed ground truth} camera poses.}
 \label{fig:fixpose_sup}
\end{figure}

\begin{figure}[t]
  \centering
\begin{minipage}[b]{0.47\hsize}\vspace{0pt}
  \centering
 \includegraphics[width=\hsize]{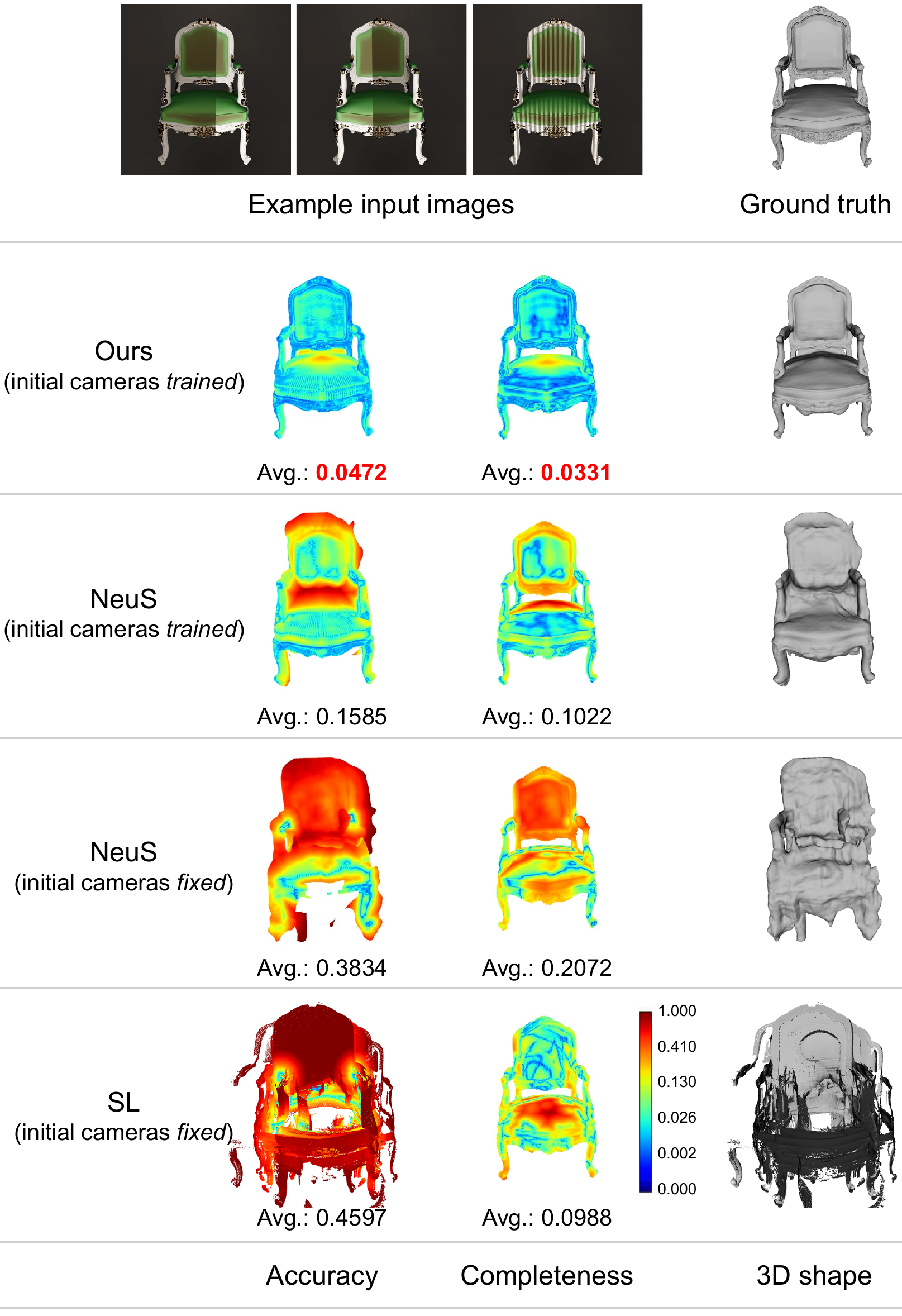}
 {\footnotesize (a) Chair}\\
\end{minipage}
\hspace{10pt}
\begin{minipage}[b]{0.47\hsize}\vspace{0pt}
  \centering
 \includegraphics[width=\hsize]{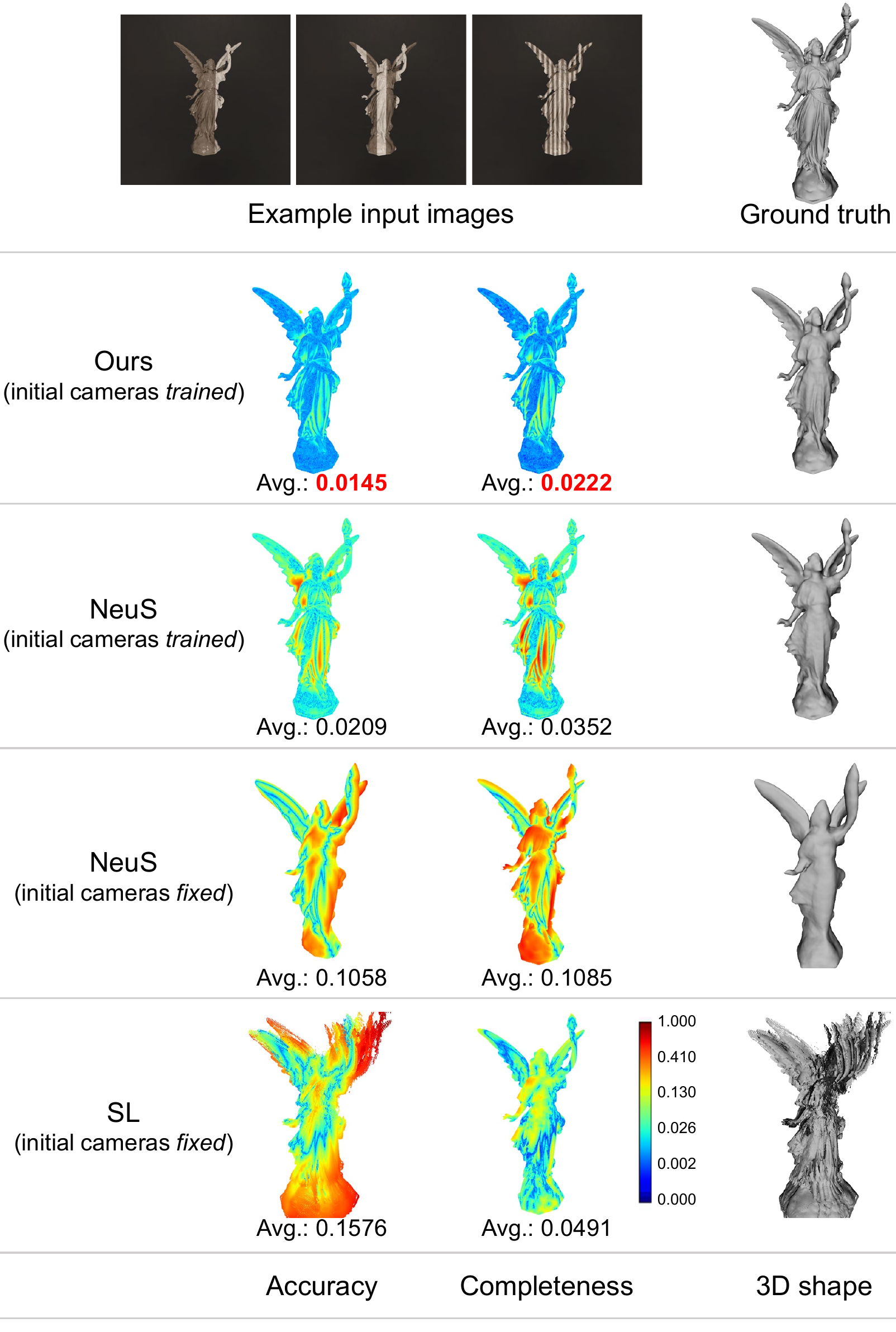}
 {\footnotesize (b) Lucy (marble)}\\
\end{minipage}
 \caption{Example input images, 3D reconstruction results, and their completeness and accuracy errors on two additional synthetic scenes with {\it noisy} camera poses.}
 \label{fig:optpose_sup}
\end{figure}

\section{Additional experimental results}

\subsection{Simulation results}
In this section, we show additional quantitative simulation results on a Stanford Bunny model (Fig.~\ref{fig:fixpose_sup} (a)), Happy Buddha model (Fig.~\ref{fig:fixpose_sup} (b)) and a Lucy model  (Fig.~\ref{fig:optpose_sup} (b)) obtained from the Stanford 3D Scanning Repository~\cite{stanford} and a Chair model with thin structure downloaded from the Internet~\cite{chair}.
To demonstrate the proposed method on the challenging targets, we rendered the models from the Stanford 3D Scanning Repository with different shiny materials, such as glass (Stanford Bunny), metal (Happy Buddha) and marble (Lucy).
For each synthetic scene, the input images are generated using the same setup as described in Section~4.1 of the main paper.
We used our method to generate 3D reconstructions in two different setups: (1) {\it fixed ground-truth} camera poses and (2) trainable camera poses with {\it noisy} initializations obtained using an SfM approach~\cite{Schoenberger}.
Fig.~\ref{fig:fixpose_sup} shows the comparisons with baseline methods with {\it fixed ground truth} camera poses.
Fig.~\ref{fig:optpose_sup} shows the comparisons with baseline methods with {\it noisy} camera poses calculated by Colmap.
In Table~\ref{table:camerapose_sup} we show a comparison of camera directions (Dire.) and positions (Posi.) between the noisy initial values and optimized values (Opt.).
Note the considerable improvement in optimized camera accuracy over initial values.

\begin{table}[t]
\small
\centering
\caption{Camera poses accuracy w.r.t the ground truth. }
    \label{table:camerapose_sup}
    \vspace{0mm}
    \small
        \begin{tabular}{lcccc}
            \toprule
            &\multicolumn{2}{c}{Chair}&\multicolumn{2}{c}{Lucy}\\
\cmidrule(lr){2-3}
\cmidrule(lr){4-5}
            &Initial & Opt. & Initial & Opt. \\
            \midrule
             Dire.(deg) & 2.832& \textbf{0.177} & 0.781 & \textbf{0.106}\\
             Posi.(m) & 0.119& \textbf{0.037}& 0.830& \textbf{0.044}\\
            \bottomrule
        \end{tabular}
\end{table}

\begin{table}[t]
\small
\centering
\caption{Quantitative results of ablation studies.} 
    \vspace{0mm}
\begin{tabular}{lcc}%{m{5cm}}
    \toprule
    & Avg. of acc. & Avg. of comp.\\
    \midrule
    (a) w/o $\mathcal{L}_{\rm SR}$ & 0.0101 & 0.0157\\
    (b) w/o $\mathcal{L}_{\rm ST}$ & 0.0114 & 0.0160\\
    (c) w/o noise reduction & 0.0174 & 0.0183\\
    (d) initial cameras fixed & 0.0191 & 0.0194\\
    (e) full model & 0.0094 & 0.0155\\
    \bottomrule
\end{tabular}
    \label{table:ablation}
\end{table}

\subsection{Ablation studies}
\label{ssec:ablationstudies}

We used the glossy marble Dragon model (the same scene in Fig.~6 of the main paper) to conduct the ablation study. 
First, to confirm the contribution of the individual loss used for structured-light supervision (reprojection loss $\mathcal{L}_{\rm SR}$ and triangulation loss $\mathcal{L}_{\rm ST}$), we test following two cases: (a) w/o $\mathcal{L}_{\rm SR}$ (by setting $\lambda_{\rm SR} = 0$), (b) w/o $\mathcal{L}_{\rm ST}$ (by setting $\lambda_{\rm ST} = 0$).
The quantitative results are shown in Table~\ref{table:ablation}.
We can confirm that the (e) full model that uses both of $\mathcal{L}_{\rm SR}$ and $\mathcal{L}_{\rm ST}$ achieves the best result.
We also studied the effect of the noise reduction of decoding. The noises caused by inter-reflection leads to a deteriorated reconstruction quality as shown in Table~\ref{table:ablation} (c) when compared with the (e) full model which reduced the noises.
In Table~\ref{table:ablation} (d) we show the result of training with fixed camera poses set to the inaccurate camera initializations obtain with SfM~\cite{Schoenberger}. This indicates that the joint optimization of camera poses and 3D geometry is indeed significant.

\begin{figure}[t]
  \centering
 \includegraphics[width=0.9\hsize]{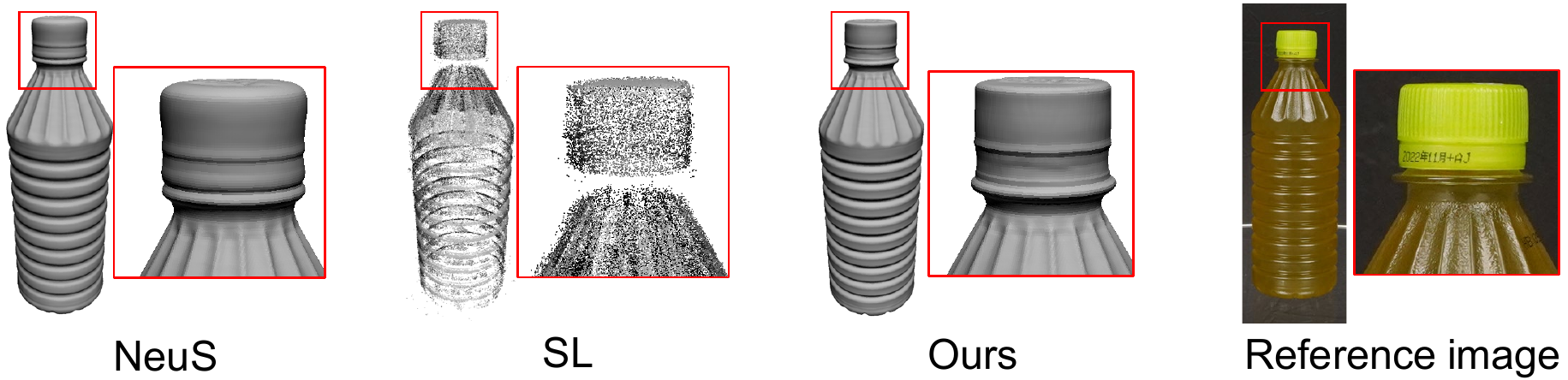}\\
 {\footnotesize (a) Plastic bottle}\\
 \vspace{12px}
 \includegraphics[width=0.9\hsize]{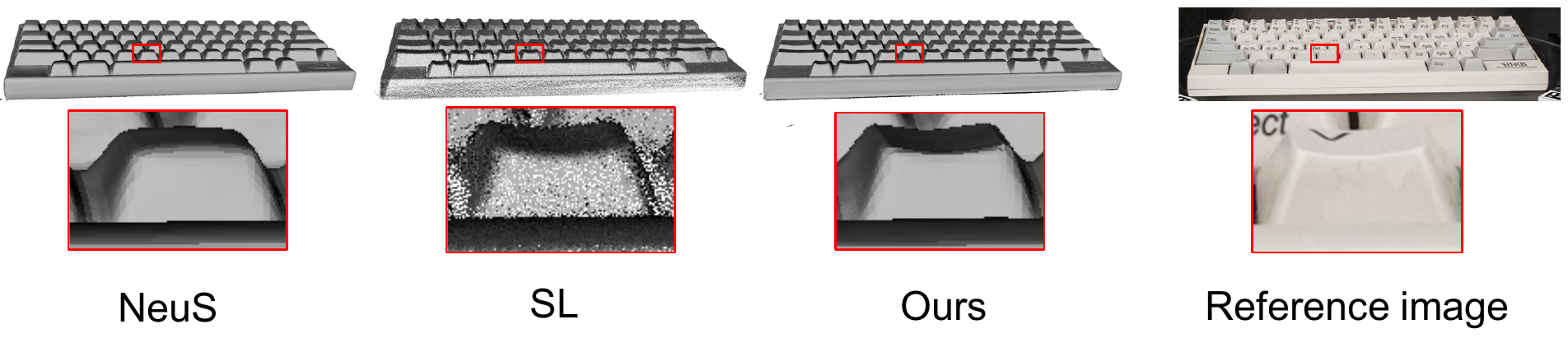}\\
 {\footnotesize (b) Keyboard}\\
 \caption{Additional 3D reconstruction results on the real dataset.}
 \label{fig:real_sup}
\end{figure}

\subsection{Results for real-world scenes}

In Fig.~\ref{fig:real_sup} we present additional qualitative results on the real dataset.
The data acquisition follows the same setup as described in Section~4.1 of main papaer.
We can confirm that proposed method perform better than all baseline methods.
%, even on the object with thin structure (Fan and Hanger).  

\subsection{Limitations}

\begin{figure}[t]
\centering
%\raisebox{0pt}[\dimexpr\height-5\baselineskip\relax]{        \includegraphics[scale=0.4]{img/limitation.pdf}}%
 \includegraphics[width=0.55\textwidth]{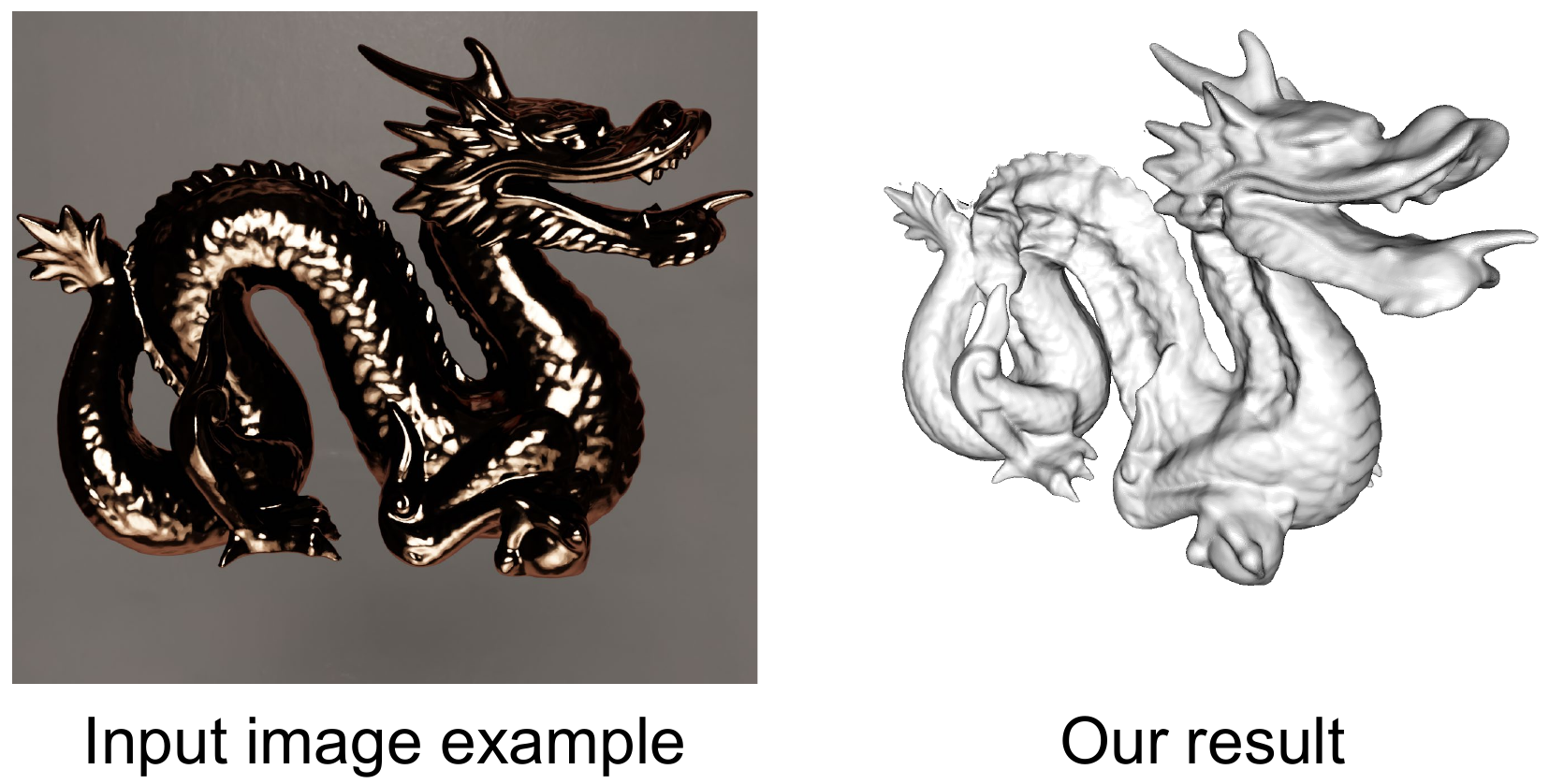}
\caption{A failure case on a mirror-like object.}
\label{fig:limitation}
\end{figure}
Although our method produces satisfactory results in most cases, it has several limitations.
First, the projector pattern will not be captured by the cameras, and no correspondences can be obtained if the material of the object is mirror-like. In this case our method only relies on photometric supervision. In Fig.~\ref{fig:limitation} we show a failure case on a synthetic scene with a textureless and mirror-like reflection. Our method fails to reconstruct an accurate surface owing to the lack of structured-light supervision. It should be noted that this material is also challenging for other state-of-the-art methods.
Second, although our method can optimize camera poses, it requires a reasonable camera pose initialization using markers or SfM softwares.

\bibliography{egbib}

%% file: s0_abstract.tex
\begin{abstract}
% Three-dimensional (3D) object reconstruction based on differentiable rendering (DR) is an active research topic in computer vision.
% However, most approaches perform poorly for textureless objects due to the ambiguity of their geometrical observations.
% In this work, we introduce active sensing with structured light (SL) into multi-view 3D object reconstruction based on differentiable rendering to learn the unknown geometry and appearance of the arbitrary scenes and the camera poses. More specifically, our framework leverages the knowledge of correspondences between cameras extracted using the structure-light pattern to optimize the implicit surface representation.
% We apply a signed distance function (SDF) and color field to represent scene geometry and appearance, respectively.
% These neural networks, together with unknown camera poses, are refined by optimizing the consistency of stereo correspondences and fidelity of differentiable rendered images. 
% In our method, while active sensing reduces geometry ambiguity in textureless regions, gradient-based optimization of consistency with differentiable rendering facilitates the calibration of camera poses, which is required for conventional SL-based methods.
% Experimental results on both synthetic and real data demonstrate that our system outperformed conventional DR and SL-based methods in high-quality surface reconstruction, especially for challenging objects with textureless, shiny, or transparent surfaces.

Three-dimensional (3D) object reconstruction based on differentiable rendering (DR) is an active research topic in computer vision.
DR-based methods minimize the difference between the rendered and target images by optimizing both the shape and appearance and realizing a high visual reproductivity.
However, most approaches perform poorly for textureless objects because of the geometrical ambiguity, which means that multiple shapes can have the same rendered result in such objects.
To overcome this problem, we introduce active sensing with structured light (SL) into multi-view 3D object reconstruction based on DR to learn the unknown geometry and appearance of arbitrary scenes and camera poses. More specifically, our framework leverages the correspondences between pixels in different views calculated by structured light as an additional constraint in the DR-based optimization of implicit surface, color representations, and camera poses. Because camera poses can be optimized simultaneously, our method realizes high reconstruction accuracy in the textureless region and reduces efforts for camera pose calibration, which is required for conventional SL-based methods.
Experiment results on both synthetic and real data demonstrate that our system outperforms conventional DR- and SL-based methods in a high-quality surface reconstruction, particularly for challenging objects with textureless or shiny surfaces.

% \keywords{3D reconstruction, multi-view stereo, structured-light, differentiable rendering, neural fields}

\end{abstract}

%% file: s1_introduction.tex
\section{Introduction}
\label{sec:introduction}

% Introduction of 3D object reconstruction
% Importing 3D objects from the real world into virtual worlds is an essential technology in the development of virtual reality (VR) and augmented reality (AR) applications. Automated 3D geometry reconstruction from 2D images aided by computer vision is known as multi-view stereo or photogrammetry. Outside AR/VR, this technique has many practical applications, such as in cultural heritage protection~\cite{LiR,Niven}, paleontology~\cite{ziegler2020applications}, visual inspection~\cite{huang2019improved}, and 3D printing~\cite{biehler20143d}.
Importing 3D objects from the real world into virtual worlds is an essential technology in digital arts, extended reality (XR) applications, cultural heritage protection~\cite{LiR,Niven}, paleontology~\cite{ziegler2020applications}, visual inspection~\cite{huang2019improved}, and 3D printing~\cite{biehler20143d}.
Conventional methods applied to achieve this automatically include photogrammetry, which integrates multiple 2D images from different views.

\begin{figure}[t]
\centering
 \includegraphics[width=0.95\hsize]{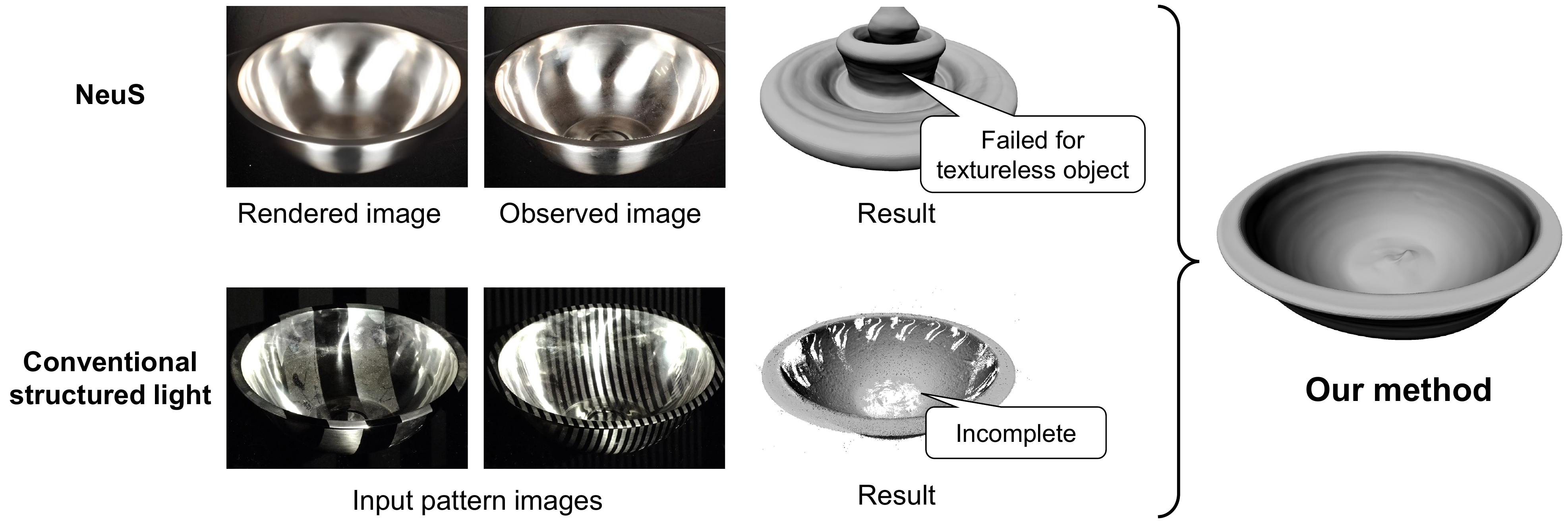}
 \vspace{-10px}
 \caption{3D reconstruction results on a textureless and shiny object (metallic bowl). NeuS~\cite{neus}, which only uses photometric information to supervise a learning process, failed to reconstruct the textureless and concave parts, even if the rendered images were close to the observed images.
 However, conventional structured-light (SL) technology, which is competent in a textureless surface, provides incomplete results owing to the highlights in captured pattern images for shiny object.
 By contrast, our method can achieve high-quality 3D reconstruction by combining SL technology and DR-based multi-view stereo.}
\label{fig:bowl}
 \vspace{-15px}
\end{figure}

% Introduction of DR-based reconstruction such as IDR
Differentiable rendering (DR)~\cite{kato2020differentiable,zhao2020physics} is an emerging tool for multi-view stereo. In contrast to traditional methods that rely on matching features from different views~\cite{camp,furu,gipuma,tola}, in DR-based methods, 3D geometries are directly optimized using gradient descent, such that rendered images are close to the observed images. A high-fidelity reconstruction can be achieved~\cite{dvr,idr} when combined with 3D representations using neural networks ({\it neural fields}~\cite{neuralfields}).
However, DR-based methods are subject to an inherent problem in terms of the geometrical ambiguity of the observations. In other words, the observed images can be explained by several different geometries, although which among them is more accurate cannot be determined. This ambiguity is high on concave or textureless surfaces, such as the result produced by NeuS, as shown in Fig.~\ref{fig:bowl}.

% Weakness of DR

% Introduction of structured light

Active vision alleviates this problem. A representative method is the use structured light (SL)~\cite{Caspi,GuQ,Nguyen,Salvi,li2019pro,Li7},  in which multiple texture patterns are projected onto objects using a projector. Although the SL system requires an additional device (the projector), it can measure the depth (3D point clouds) with high quality even for textureless objects owing to an active projection. However, standard SL systems have certain disadvantages. First, the camera and projector must be rigidly fixed (mounted on a rig) and precisely calibrated. second, the SL system provides a 3D reconstruction of poor quality to recover optically complex scenes such as shiny objects, because the appearance of highlights and inter-reflections lead to incomplete 3D point cloud result. Third, SL systems are sensitive to occlusions, and by extension to holes, because they rely on an optical disparity.

% Proposal
In this work, we propose to introduce active sensing with SL into DR-based multi-view stereo.
We follow the implicit differentiable renderer~\cite{idr,neus,volsdf} to represent the surface as a zero-level set of a signed distance field (SDF) and the scene appearance as a color field.
To train these networks, a sequence of SL patterns is projected onto the object surface and captured by two cameras while the object is arbitrarily rotated to obtain multiple observations of SL patterns around the object (see Fig.~\ref{fig:configuration}). The correspondence between the two camera views extracted from the captured images with projected patterns is used to supervise the training of the object surface.
In addition, images without a SL pattern for each camera view are captured for photometric supervision, which has been widely used in previous DR frameworks. Both the surface and appearance of the object are optimized by minimizing the loss between observations and rendered images.
On the one hand, compared to previous DR frameworks~\cite{idr,neus,volsdf} which only use photometric data to supervise the training, the SL supervision can provide important cues to reduce the geometrical ambiguity. 
On the other hand, shiny objects that cannot be handled by SL technology can be approximately represented using the color field. Thus, their shapes can be optimized through photometric supervision.
In addition, self-occlusion, which cannot be handled by SL technology, can be solved through the global optimization of the surface represented by an implicit SDF using photometric supervision.
Furthermore, by establishing a geometric relationship between the neural implicit surface and camera poses, the camera extrinsic parameters are also refined during optimization; therefore, our method does not require strict calibration in contrast to conventional SL methods~\cite{Caspi,GuQ,Nguyen,Salvi}.
To the best of our knowledge, our work is the first to combine SL and differentiable rendering.

% Experiment
We experimentally validated the effectiveness of the proposed method under both synthetic and real-world scenarios.
The results demonstrate that our method can reconstruct many types of challenging targets that are textureless, self-occluded, or shiny even with rough camera information. They also show that the proposed approach outperforms other state-of-the-art neural scene representation methods as well as conventional SL methods.

%% file: s3_method.tex
 \vspace{-0.5em}
\section{Method}

We aim to reconstruct the 3D shape of an object from multi-view structured-light (SL) pattern images with rough camera poses with known intrinsic parameters. Additionally, the proposed method does not require mask supervision.
Inspired by IDR~\cite{idr}, NeuS~\cite{neus}, and VolSDF~\cite{volsdf}, we adopt a neural implicit SDF and use a zero-level set to represent the surface of the object. To obtain a high-quality 3D model result, we introduce SL pattern consistency to supervise the neural network training.
Our approach combines SL and DR-based multi-view stereo to ensure self-calibration and high-quality 3D reconstruction of objects.

\begin{figure}[!tbp]
  \centering
 \includegraphics[width=\hsize]{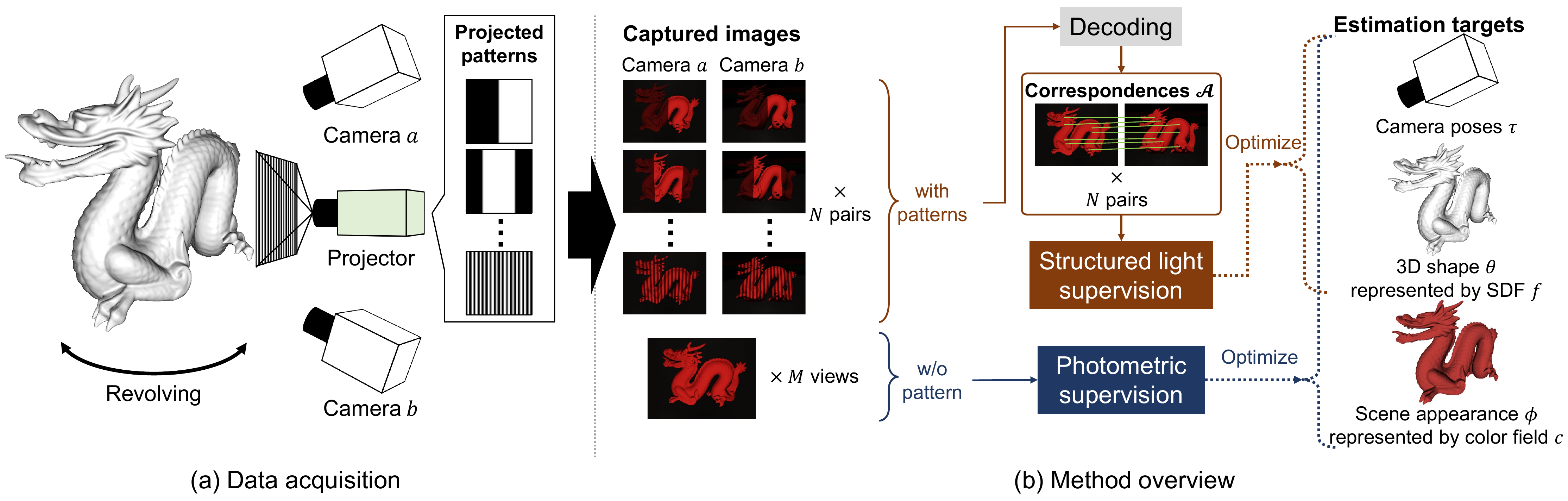}
 \vspace{-20px}
  \caption{We use the projector to project a sequence of gray code patterns. To scan the whole shape of the object, we rotate it repeatedly until the whole shape of the object is scanned. We obtain $N$ pairs of image sequences with projector patterns (used in SL supervision) and $M$ multi-view images without projector pattern (used in photometric supervision).}
  \label{fig:configuration}
 \vspace{-12px}
\end{figure}

\vspace{-1em}
\subsection{Data acquisition and pattern decoding}
\label{ssec:dataacquisition}
 \vspace{-0.5em}

\noindent {\bf Data acquisition:}
Fig.~\ref{fig:configuration} (a) illustrates the data-acquisition procedure of proposed method.
A projector was used to project a sequence of SL patterns, and two cameras ($a$ and $b$) were used as imaging devices to capture images with projector patterns (used in SL supervision) and an image without the pattern (used in photometric supervision).
To scan the entire shape of the object, we changed the viewpoint of the cameras and projector by rotating the object, as shown in Fig.~\ref{fig:configuration} (a).
Finally, we obtain $N$ pairs of images sequences with projector patterns (used in SL supervision) and $M$ multi-view images without projector pattern (used in photometric supervision).

Images with patterns were further used to perform SL supervision. In our experiment, we generated patterns by encoding each pixel coordinate $\bm{q}$ of the projector to a binary gray code~\cite{Geng,GuQ}.
In contrast to other patterns~\cite{heike20103d,payeur2009structured,pribanic2010efficient} that use different levels of intensity or different color to produce unique coding, gray code pattern only uses binary values (white and black), thus it is less sensitive to the surface characteristics.
Our system generates $\log_{2}O$ bits of gray code to assign an independent value to each pixel, where $O$ is the number of projector rows (or columns). To improve the decoding accuracy, similar to conventional SL methods~\cite{GuQ}, the gray code patterns insert the original and its inverse to identify code words, a white image, and black illumination to identify shadow regions.
Therefore, a sequence of gray code patterns consists of 46 frames for a projector with the resolution of 1920 $\times$ 1080 (including $2 \times \lceil\log_{2}1920\rceil=22$ patterns representing the columns, $2 \times \lceil\log_{2}1080\rceil=22$ patterns representing the rows, and one pair of white and black images).

\vspace{1mm}
\noindent {\bf Pattern decoding and noise reduction:}
After capturing the images we follow the decoding algorithm in~\cite{GuQ} to decode each pixel $\bm{p}$ in the images into their corresponding decimal number $\bm{q}$ representing the column and the row of the projector.
However, some noise will be captured due to the inter-reflection of the projected pattern, especially for shiny or concave surfaces.
Therefore we determine whether a decoded pixel is affected by inter-reflection using the epipolar line. As the camera poses are unknown in our experiment, we calculate a rough fundamental matrix between the camera and projector from the noisy corresponding points, and estimate the epipolar lines using this fundamental matrix.
Then, we eliminate correspondences whose camera pixels are not on the epipolar line.
More details about noise reduction are in the supplementary material.
% As shown in Fig.~\ref{fig:noisereduction}, the light projected from the projector pixel $\bm{q}$ can reach the camera in one of two general ways: (1) by direct surface reflection, captured by a camera pixel $\bm{p}$ on the epipolar line (black path), which is the desirable path of the light for pattern decoding, or (2) by inter-reflection, captured by a camera pixel $\bm{p}^{\prime}$ that is not on the epipolar line (orange path).
% Therefore, we can determine whether a decoded pixel is affected by inter-reflection using the epipolar line.
% As the camera poses are unknown in our experiment, we calculate a rough fundamental matrix between the camera and projector from the noisy corresponding points and estimate the epipolar lines using this fundamental matrix.
% Then, we eliminate correspondences whose camera pixels are not on the epipolar line.
% Note that although we can effectively reduce most noise using this strategy, some limitations remain: (1) the estimated epipolar lines may include minor errors owing to the noisy corresponding points, and (2) we cannot eliminate the inter-reflected correspondences whose projector and camera pixels are on corresponding epipolar lines.
% However, the amount of noise caused by these cases is small, so they can be further reduced by the photometric supervision introduced in Section~\ref{ssec:photometricsupervision}.
% The effectiveness of this noise-reduction strategy is demonstrated by the results of our experiments (see Section~\ref{ssec:ablationstudies}).

Finally, for each camera pair, we can map the pixels that share the same corresponding projector pixel $\bm{q}$ to output a list $\mathcal{A}$ of \{$\bm{p}_{a}$, $\bm{p}_{b}$\}, where $\bm{p}_{a}$ and $\bm{p}_{b}$ denote the pixel coordinate in camera $a$ and $b$, respectively.
By repeating the process for all $N$ camera pairs, we can obtain $N$ corresponding lists.

 \vspace{-1em}
\subsection{Geometry, appearance and camera pose representations}
 \vspace{-0.5em}
In our network, the surface $S(\theta)$ is modeled explicitly as the zero-level set of an SDF, which is represented by an MLP $f(\bm{x};\theta): \mathbb{R}^3 \rightarrow \mathbb{R}$ with learnable parameters $\theta$. The network $f$ takes a query location $\bm{x}\in \mathbb{R}^3$ as input and outputs a signed distance from the location to the closest surface point
%, with the sign determined by whether $\bm{x}$ is on the surface $S(\theta)$
(a positive distance for $\bm{x}$ outside and a negative distance for $\bm{x}$ inside). $S(\theta)$ can be represented as
\[
S(\theta)= \left\{ \bm{x} \in \mathbb{R}^3 \mid f(\bm{x};\theta) = 0\right\}.
\label{eq:geometry}
\]

We denote the camera poses (extrinsic parameters), including the camera positions and rotations, using the parameter $\tau$, which are also optimized during network training. We assume the intrinsic parameters of the cameras are known. Given a pixel in a camera view, let $R(\tau)$ denote the ray through this pixel, and we obtain 
\begin{equation}
R(\tau) = \left\{ \bm{o} + t\bm{v} \mid t \geq 0 \right\},
\label{eq:ray}
\end{equation}
where $\bm{o}=\bm{o}(\tau) \in \mathbb{R}^3$ is the center of the camera, $\bm{v}=\bm{v}(\tau) \in \mathbb{R}^3$ is the unit direction vector of the ray, and $t$ is the distance from point $\bm{x}$ to camera center $\bm{o}$. Here, we suppose that the surface exists only in an interval $\lbrack t_{\rm L}, t_{\rm R}\rbrack$ such that $t \in \lbrack t_{\rm L}, t_{\rm R}\rbrack$.

The scene appearance is represented by a color field using another MLP $c(\bm{x},\bm{v};\phi): \mathbb{R}^3 \times \mathbb{R}^3 \rightarrow \mathbb{R}^3$ with learnable parameters $\phi$. This MLP $c$ encodes the RGB color associated with the query location $\bm{x}\in \mathbb{R}^3$ and the viewing direction $\bm{v}\in \mathbb{R}^3$.

% \begin{figure}[!tbp]
% \begin{minipage}{0.38\hsize}
% \centering
% \includegraphics[width=\textwidth]{img/noisereduction.pdf}
% %  \vspace{-8px}
% \caption{Pattern misdetection caused by inter-reflection.}
% \label{fig:noisereduction}
% \end{minipage}
% \begin{minipage}{0.60\hsize}
%   \centering
%  \includegraphics[width=\hsize]{img/slsupervision.pdf}
% %  \vspace{-8px}
%   \caption{Calculation of reprojection loss $\mathcal{L}_{\rm SR}(\theta, \tau)$ (left) and triangulation loss  $\mathcal{L}_{\rm ST}(\theta, \tau)$ (right).}
%   \label{fig:slsupervision}
% \end{minipage}
%  \vspace{-12px}
% \end{figure}

\begin{wrapfigure}[12]{R}[0pt]{0pt}
\centering
 \includegraphics[width=0.49\textwidth]{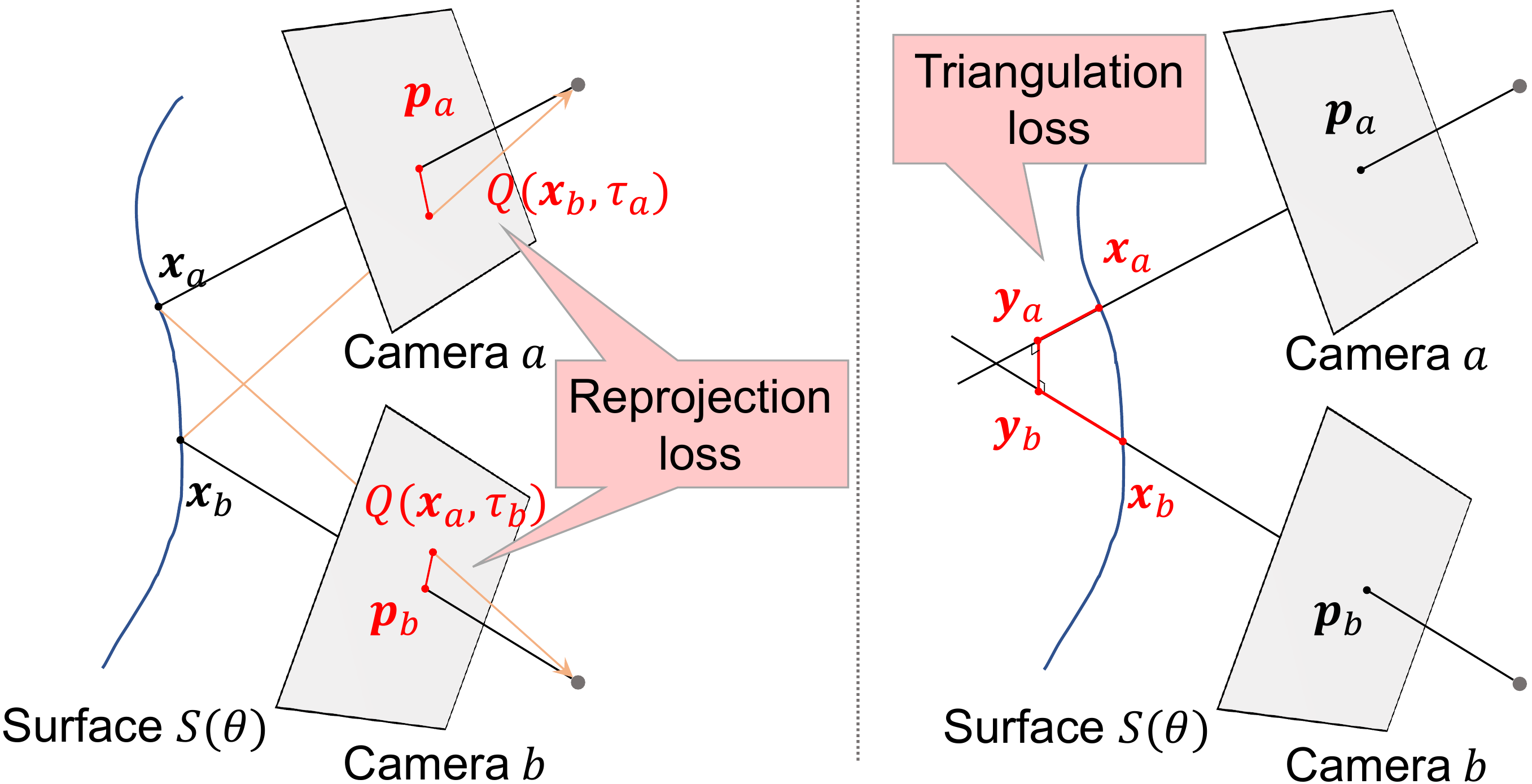}
  \vspace{-20pt}
\caption{Reprojection loss $\mathcal{L}_{\rm SR}(\theta, \tau)$ (left) and triangulation loss $\mathcal{L}_{\rm ST}(\theta, \tau)$ (right).}
\label{fig:slsupervision}
\end{wrapfigure}

\vspace{-1em}
\subsection{SL supervision}
\vspace{-0.5em}
\label{ssec:structuredlight}
For high-quality 3D shape reconstruction, our main idea is to exploit the dense and accurate correspondences extracted by SL patterns as constraints during the optimization of the 3D shapes and camera poses.

Given the extracted correspondences between each camera pair, we calculate the intersections between the surface and the ray passing through each pixel, as shown in Fig.~\ref{fig:slsupervision}.
%%%% Is this part necessary?
Following previous works~\cite{dvr,idr}, we first use a ray marching algorithm to find the intersection point, and construct differentiable intersection which has a correct value and first-order derivative with respect to $\theta$ and $\tau$.
 Let $\bm{x} = \bm{o} + t\bm{v}$ denote the intersection point of the ray $R(\tau)$. For the current network parameters $\theta_0$ and camera parameters $\tau_0$, we denote $\bm{o}_0=\bm{o}(\tau_0)$, $t_0=t(\theta_0,\tau_0)$, $\bm{v}_0=\bm{v}(\tau_0)$, and $\bm{x}_0 = \bm{o}_0 + t_0\bm{v}_0$. We take the implicit differentiation of equation $f(\bm{x}; \theta) \equiv 0$, and the surface intersection is expressed as a function of $\theta$ and $\tau$:
\begin{equation}
\bm{x}(\theta,\tau)=\bm{o} + t_{0}\bm{v} - \frac{f(\bm{o} + t_{0}\bm{v};\theta)}{\nabla_{\bm{x}}f(\bm{x}_0;\theta_0) \cdot \bm{v}_0}\bm{v},
\label{eq:intersection}
\end{equation}
where $\nabla_{\bm{x}}f(\bm{x}_0;\theta_0)$ is constant.
%%%%%

We consider two types of pattern consistency: reprojection loss and triangulation loss. From our experiment (see supplementary material), we find that using both loss functions results in better reconstruction results than using only one of them.

% \subsubsection{Reprojection loss:}
\vspace{1mm}
\noindent {\bf Reprojection loss:}
The reprojection loss ensures that a surface point $\bm{x}$ on a ray is projected near pixel $\bm{p}$ from another view that corresponds to the ray, as shown in Fig.~\ref{fig:slsupervision} left column. Concretely, it models the error between the reprojected pixels from the surface intersections and the corresponding pixels. The loss function is described as
\begin{equation}
\mathcal{L}_{\rm SR}(\theta, \tau) = \sum_n^{N} \sum_{i\in\mathcal{A}} \left(\|Q(\bm{x}_{a}^{n,i}, \tau_{b}^{n}) - \bm{p}_{b}^{n,i}\| + \|Q(\bm{x}_{b}^{n,i}, \tau_{a}^{n}) - \bm{p}_{a}^{n,i}\|\right),
\label{eq:reprojectionloss}
\end{equation}
where $Q(\bm{x}, \tau)$ is the reprojection of surface point $\bm{x}$ on the image with camera parameters $\tau$.

% \subsubsection{Triangulation loss:}
\vspace{1mm}
\noindent {\bf Triangulation loss:}
The triangulation loss enforces consistency between the estimated 3D shape and 3D point cloud, which is directly calculated by triangulation from the extracted correspondences between each camera pair.
As shown in the right column of Fig.~\ref{fig:slsupervision}, suppose we obtain a correspondence between camera pair $a$ and $b$ by decoding the SL pattern. We can separately calculate the intersections $\bm{x}_a$ and $\bm{x}_b$ between the surface and the two camera rays via ray tracing as described above.
In addition, we can obtain the intersection between these two camera rays by triangulation. However, considering that these two rays may not intersect because of the camera pose error, we calculate the closest point $\bm{y}_a$ to the ray $R_b(\tau)$ on the ray $R_a(\tau)$ and the closest point $\bm{y}_b$ to the ray $R_a(\tau)$ on the ray $R_b(\tau)$. More details on the calculation of $\bm{y}_a$ and $\bm{y}_b$ are provided in the supplementary material.
% \begin{equation}
% \bm{x}^{tria}_{a} = \frac{(\bm{o_b}-\bm{o_a})\cdot\bm{v_a} - }{}
% \label{eq:triangulation}
% \end{equation}
Finally, to ensure that these four points ($\bm{x}^{n,i}_a$, $\bm{x}^{n,i}_b$, $\bm{y}^{n,i}_a$ and $\bm{y}^{n,i}_b$) are located in the same position, we calculate the triangulation loss which evaluates the distance between these four points, as given below.
\begin{equation}
\mathcal{L}_{\rm ST}(\theta, \tau) = \sum_n^{N} \sum_{i\in\mathcal{A}} \left(\|\bm{x}^{n,i}_a - \bm{y}^{n,i}_a\| + \|\bm{y}^{n,i}_a- \bm{y}^{n,i}_b\| + \|\bm{y}^{n,i}_b - \bm{x}^{n,i}_b\|\right).
\label{eq:triangulationloss}
\end{equation}

\vspace{-1em}
\subsection{Photometric supervision}
\vspace{-0.5em}
\label{ssec:photometricsupervision}
The SL supervision in Section~\ref{ssec:structuredlight} can correctly recover the surface geometry.
However, as the correspondences extracted using SL patterns are usually noisy and incomplete for some special materials, such as shiny surfaces, we propose to consider rendered image consistency during network training.

Because we acquire multi-view observations by rotating an object with fixed camera positions, as described in Section~\ref{ssec:dataacquisition}, the background for each observation is constant and can be easily obtained in advance. Therefore, our SDF $f$ and color field $c$ only represent the shape and appearance of the foreground (i.e., the object). The rendered images are obtained by mixing the rendered foreground images from the neural networks and known background images.
For foreground rendering, we use the same rendering method as NeuS~\cite{neus}, which adopts a volume-rendering scheme.
Specifically, the output color for pixel $k$ in the rendered image is calculated by accumulating the weighted colors along the ray.
Because the MLP would only be queried at a discrete set of locations, the viewing ray is sampled by partitioning $\lbrack t_{\rm L}, t_{\rm R}\rbrack$ into $n$ evenly-spaced bins, and Eq.~(\ref{eq:ray}) can be rewritten as follows.
\begin{equation}
R(\tau) = \left\{ \bm{o} + t_j\bm{v} \mid j = 1, ..., n, t_j<t_{j+1}\right\}.
\label{eq:raysampling}
\end{equation}
The foreground rendering formula for this ray can be defined as
\begin{equation}
C_{\rm fore}(\theta, \phi, \tau)=\sum_{j=1}^{n}w_j(\theta) c(\bm{o} + t_j\bm{v},\bm{v}; \phi),
\label{eq:foregroundrendering}
\end{equation}
where $w_j(\theta)$ is the weight of the sampled point $\bm{o} + t_j\bm{v}$, which is a function of the distance to the surface $S(\theta)$. Thus, the weight is the connection between the output colors and the implicit SDF $f$. For details regarding the weight function, please refer to~\cite{neus}.
The rendered color is then calculated by mixing the estimated foreground $C_{\rm fore}(\theta, \phi, \tau)$ and the known background $C_{\rm back}$.
\begin{equation}
C(\theta, \phi, \tau)=C_{\rm fore}(\theta, \phi, \tau) + C_{\rm back}\left(1 - \sum_{j=1}^{n}w_j(\theta)\right).
\label{eq:rendering}
\end{equation}

Even though our method assumes known background images, in our setup obtaining background images is easier than calculating accurate object masks. 
The object masks have to be generated from input images by manual annotation or some automatic foreground detection methods. 
However automatic methods are sometimes inaccurate, so eventually the manual annotation might be required.
And the performance of conventional methods highly relies on the accuracy of the object masks. 
Thus the mask generation is always a costly process for conventional methods.
On the other hand, our method uses a turntable to change the direction of the object, thus the relative positions between cameras and the background are fixed (i.e. the background is constant in all viewpoints for one camera).
Therefore, we just capture one background image (without object) for each camera before (or after) object capture.

Finally, given the rendered color $C^k = C^k(\theta, \phi, \tau)$ of pixel $k$ and the input image color $I^k$ of pixel $k$, the render loss is calculated as the L1 distance.
\begin{equation}
\mathcal{L}_{\rm R}(\theta, \phi, \tau) = \sum_{k\in\mathcal{V}_I} | I^k-C^k |,
\label{eq:renderingloss}
\end{equation}
where $\mathcal{V}_I$ denotes the set of all image pixels in all camera views without patterns.

For shiny surfaces, the 3D point clouds extracted from SL patterns are usually incomplete, as mentioned above. Thus we consider the photometric supervision by minimizing the error between the observed images and the rendered images to guarantee geometry accuracy in the missing areas of SL supervision.
However, the view-dependent (specular) reflection model of shiny surfaces using the traditional point cloud or mesh is high-complexity.
By contrast, the SDF and color field are able to efficiently represent the view-dependent scene. Thus our optimization framework introducing the DR represented by SDF is reasonable and effective to make up the shortcomings of SL methods.

\vspace{-1em}
\subsection{Training}
\vspace{-0.5em}
Similar to previous works~\cite{neus,idr}, we regularize our SDF network with an Eikonal loss function~\cite{gropp2020implicit} that restricts the expectation of the gradient magnitude to 1, as given below.
\begin{equation}
\mathcal{L}_{\rm E}(\theta) = \mathbb{E}_{\bm{x}}\left( \|\nabla_{\bm{x}}f(\bm{x};\theta)\|-1\right)^2.
\label{eq:eikonalloss}
\end{equation}
The final loss is expressed as a weighted sum of all the losses listed above.
\begin{equation}
\mathcal{L}(\theta, \phi, \tau)= \mathcal{L}_{\rm R}(\theta, \phi, \tau) + \lambda_{\rm SR}\mathcal{L}_{\rm SR}(\theta, \tau) + \lambda_{\rm ST}\mathcal{L}_{\rm ST}(\theta, \tau) + \lambda_{\rm E}\mathcal{L}_{\rm E}(\theta).
\label{eq:loss}
\end{equation}

%% file: s4_experiments.tex
\vspace{-1em}
\section{Experiments}

\vspace{-0.5em}
\subsection{Experimental setting}
\vspace{-0.5em}

% \subsubsection{Datasets.}
\noindent {\bf Datasets:}
We experimentally validated the effectiveness of the proposed method for real-world and synthetic scenes, with a wide variety of materials, appearances and geometries, including challenging cases for reconstruction algorithms, such as textureless and glossy surfaces.
Because there is no existing multi-view structured light dataset which is directly applicable to our setup, all the datasets used in our experiment were produced by the authors.

For experiments on real-world scenes, we used a Mitsubishi LVP-FD630 projector and two Sony $\alpha$6600 digital cameras in our projector-camera system. The sequence of structured light patterns was encoded with a resolution of 1920$\times$1080 and captured by the cameras using a video format with a resolution of 3840$\times$2160. We used a turntable to rotate the object, and each scene was captured from $N=12$ pairs of camera viewpoints with structured-light patterns, and $M=270$ single images without structured-light patterns.
The initial camera poses were measured using AprilTag 16h5 Markers~\cite{richardson2013iros} fixed on a turntable. Details of the strategy for the initial camera pose estimation are provided in the supplementary material.

For experiments on synthetic scenes, we used a Dragon model obtained from the Stanford 3D Scanning Repository~\cite{stanford} and a Bowl model downloaded from the Internet~\cite{turbosquid}.
To demonstrate the proposed method on the challenging targets, we rendered all the models with shiny materials, such as plastic (Fig.~\ref{fig:fixpose}, left), ceramic (Fig.~\ref{fig:fixpose}, right) and marble (Fig.~\ref{fig:trainedpose});
for each synthetic scene, we generated $N=20$ pairs of camera viewpoints with structured-light patterns and $M=40$ single images without structured-light patterns.
The evaluation of the other models is included in the supplementary material.

% \subsubsection{Evaluation metrics.}
\vspace{1mm}
\noindent {\bf Evaluation metrics:}
To evaluate the 3D shape quality, we used the accuracy (Accu.) and completeness (Comp.) as two metrics~\cite{aanaes2016large,ley2016syb3r}. The accuracy is the distance from each estimated 3D point to its nearest ground-truth 3D point. The completeness is the distance from each ground-truth 3D point to its nearest estimated 3D point.
We define the overall score, the average of mean accuracy and mean completeness, as the reconstruction quality.

% \subsubsection{Implementation details.}
\vspace{1mm}
\noindent {\bf Implementation details:}
For the MLPs of the implicit SDF $f$ and color field $c$, we followed the architectures used in IDR~\cite{idr} and NeuS~\cite{neus}.
We implemented the proposed method in PyTorch and trained our model using the Adam optimizer~\cite{kingma2014adam}.
The learning rates were first linearly warmed up from 0 to maximums ($5.0\times 10^{-4}$ for MLP training and $1.0\times 10^{-5}$ for camera pose training) in the first 5k iterations, and then controlled by the cosine decay schedule to the minimum learning rates ($2.5\times 10^{-5}$ for MLP training and $5.0\times 10^{-7}$ for camera pose training).
The loss weights in Eq.~(\ref{eq:loss}) were empirically set as $\lambda_{\rm SR}=1.0\times 10^{-4}$, $\lambda_{\rm ST}=0.1$, and $\lambda_{\rm E}=0.1$.
We sampled 512 rays per batch and trained our model for 100k iterations for 7 h with a single NVIDIA V100 Tensor Core GPU.

\vspace{-1em}
\subsection{Effectiveness of DR and SL combination}
\vspace{-0.5em}

The core of our proposed system is to combine the differentiable rendering (DR)- and structured-light (SL)-based methods to obtain the benefits of both. In this subsection, we describe our evaluation of the proposed approach.
We used our method to generate 3D reconstructions in two different setups: (1) fixed ground-truth cameras and (2) trainable cameras with noisy initializations obtained using an SfM approach~\cite{Schoenberger}.
We compared our method with a DR-based method called NeuS~\cite{neus} and an existing SL method~\cite{GuQ}.
In the SL method~\cite{GuQ}, we first reconstructed the 3D point cloud for each camera pair based on triangulation using the extracted corresponding points. We tried both the ground truth and noisy camera poses for the reconstruction. Subsequently, the 3D point cloud parts were aligned using the input camera pose.
For a fair comparison, we show the results of the existing SL method~\cite{GuQ} after the noise reduction which is described in Section~\ref{ssec:dataacquisition}, even though the original approach does not include this process.

%%%%%%%%%%%%%%%%%%%%%%%%%%%%%%%%%%%%%%%%
\begin{figure}[t]
  \centering
 \includegraphics[width=\hsize]{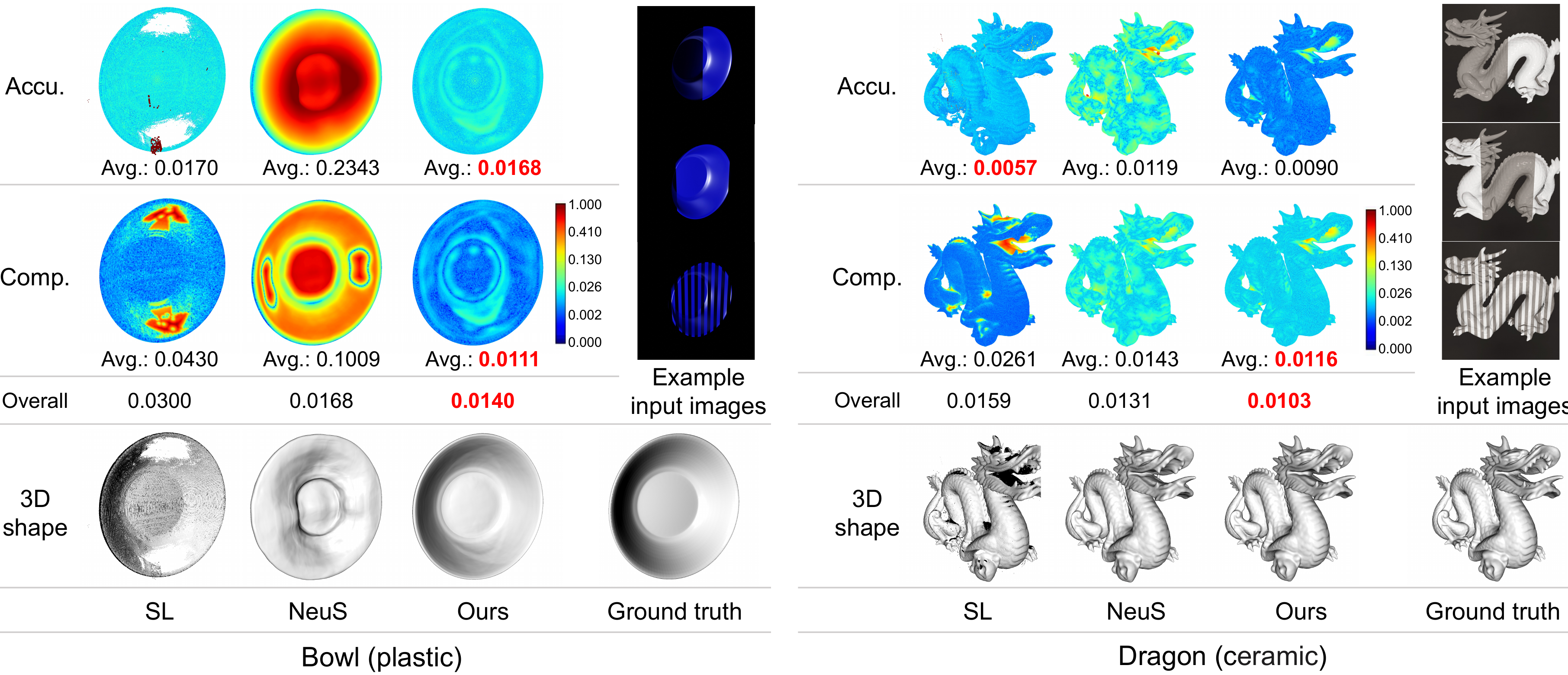}
 \vspace{-20px}
 \caption{Example input images, 3D reconstruction results, and their accuracy and completeness errors on two synthetic scenes with {\it fixed ground truth} camera poses.}
 \label{fig:fixpose}
 \vspace{-4px}
\end{figure}
%%%%%%%%%%%%%%%%%%%%%%%%%%%%%%%%%%%%%%%%

%%%%%%%%%%%%%%%%%%%%%%%%%%%%%%%%%%%%%%%%
\begin{figure}[t]
  \centering
 \includegraphics[width=\hsize]{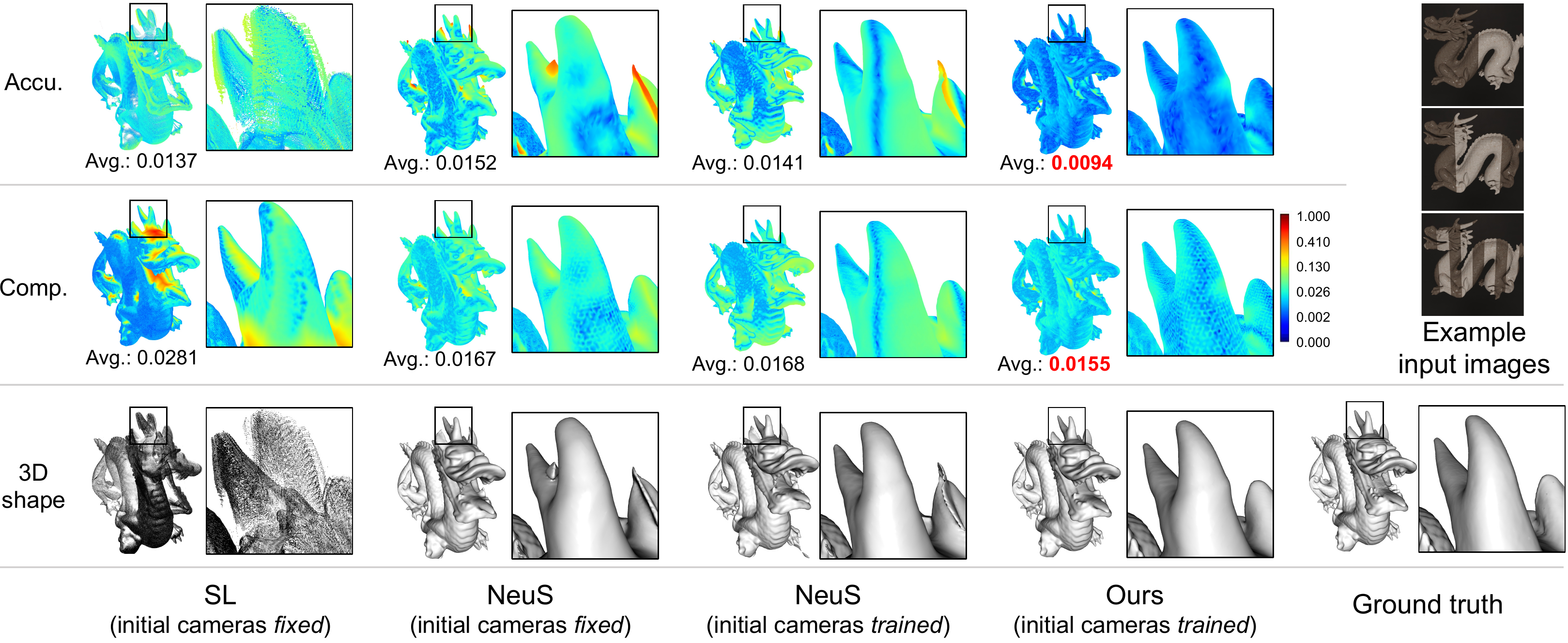}
 \vspace{-20px}
 \caption{Example input images, 3D reconstruction results, and their accuracy and completeness errors on the synthetic Dragon model (marble) with {\it noisy} camera poses.}
 \label{fig:trainedpose}
 \vspace{-12px}
\end{figure}
%%%%%%%%%%%%%%%%%%%%%%%%%%%%%%%%%%%%%%%%

Fig.~\ref{fig:fixpose} shows the reconstructed 3D shapes and their quantitative evaluations of the synthetic dataset using fixed ground-truth camera poses.
For quantitative evaluations, the calculation of accuracy and completeness is conducted using the same density of 3D point clouds.
The completeness and accuracy errors for each 3D point were colorized, and the average errors are shown below the error maps.
Compared to NeuS, which only uses photometric supervision based on a passive illumination, our method reconstructed more accurate results because structured light can reduce the geometric ambiguity of textureless surfaces.
In the SL method, the inter-reflection of the inner surface of the Bowl model results in holes, whereas the proposed method provides a complete and accurate result using photometric supervision based on differentiable rendering.
Although the SL method provides better accuracy on the Dragon model (after the noise reduction), the result of the 3D point cloud is incomplete owing to the occlusion (e.g. inside of the mouse of Dragon).
Regarding the average of accuracy and completeness results, our method was able to provide better results than the SL method.

\begin{wraptable}[8]{r}{45mm}
    \caption{Camera poses accuracy on the Dragon model (marble).}
    \small
    \begin{center}
        \begin{tabular}{lcc}
            \toprule
            & Initial & Opt. \\
            \midrule
             Dire.(deg)& 0.070 & \textbf{0.049}\\
             Posi.(m)& 0.075 & \textbf{0.011}\\
            \bottomrule
        \end{tabular}
    \end{center}
    \label{table:camerapose}
    \vspace{0mm}
\end{wraptable}

The results obtained using the initial noisy camera poses shown in Fig.~\ref{fig:trainedpose} demonstrate the effectiveness of our global optimization of the camera poses.
Although NeuS does not optimize the camera poses, we attempted to incorporate camera training into their original method for comparison.
It can be observed from the results that our method outperformed all baselines by training the camera poses using structured light supervision.
The quality of the 3D reconstruction degraded for the NeuS and SL methods, which require high-accuracy camera calibration.
In addition, our method performed better than NeuS with camera training, indicating that structured light supervision contributed to the improvement of accuracy during training for both 3D shape and camera poses.
Table~\ref{table:camerapose} shows a comparison of camera directions (Dire.) and positions (Posi.) between initial values and optimized values (Opt.).
Note the considerable improvement in optimized camera accuracy over initial values.

\vspace{-1em}
\subsection{Evaluation on real-world dataset}
\vspace{-0.5em}

%%%%%%%%%%%%%%%%%%%%%%%%%%%%%%%%%%%%%%%%
\begin{figure}[t]
  \centering
 \includegraphics[width=\hsize]{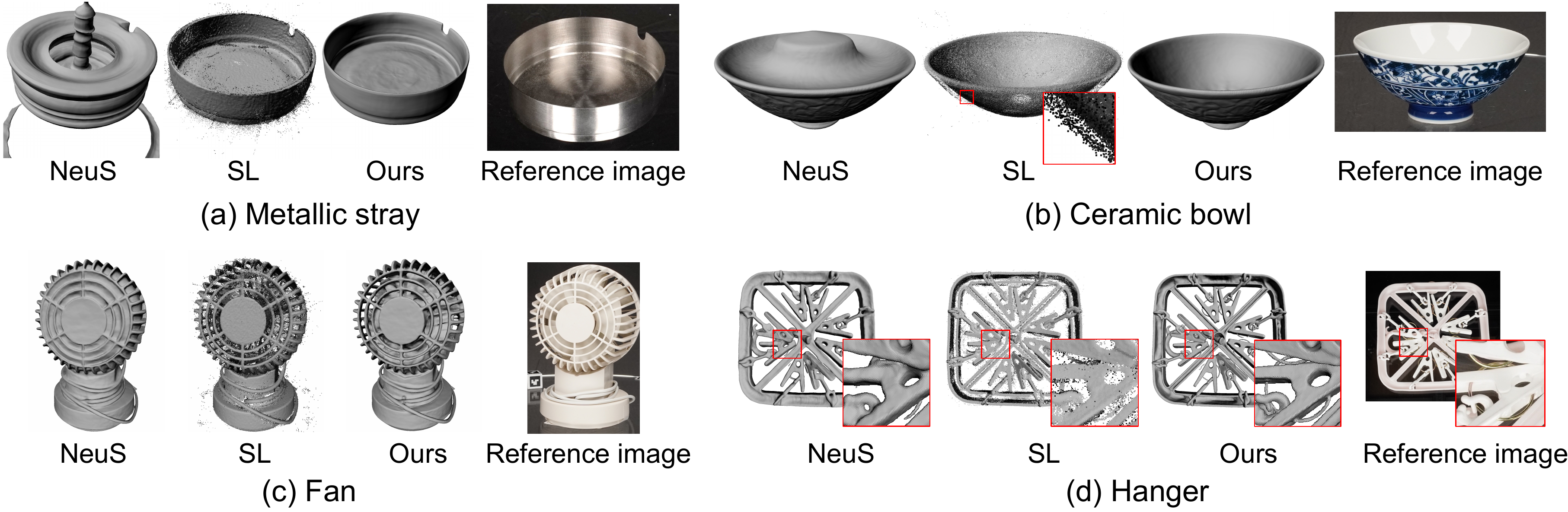}
 \vspace{-20px}
 \caption{3D reconstruction results on the real dataset.}
 \label{fig:real}
 \vspace{-12px}
\end{figure}
%%%%%%%%%%%%%%%%%%%%%%%%%%%%%%%%%%%%%%%%

So far, evaluations have been conducted on a synthetic dataset because the ground truth geometries for a quantitative evaluation are available. Here, we present the results on a real-world dataset with noisy camera parameters.
Fig.~\ref{fig:real} shows the reconstruction results for (a) a metallic ashtray, (b) a ceramic bowl, (c) a fan and (d) a hanger, compared with the NeuS and SL methods.
Both targets of the ashtray and bowl were textureless objects, which led to a geometric ambiguity for NeuS. Therefore, we can confirm that NeuS fails to reconstruct the concave parts.
In the close-up of the bowl, we can see that the SL method produces a double-layer surface. This occurred because the point clouds reconstructed from different camera pairs could not be tightly aligned based on noisy camera poses. In addition, it may be observed that the shiny surface of the ash tray led to many error points around the surface for the SL method. By contrast, our proposed approach, which benefits from both DR- and SL-based methods, provided the most accurate 3D reconstruction results.

{\bf Ablation study} and {\bf limitations} of proposed method are described in supplementary.
% \subsection{Limitations}
% \setlength\intextsep{0pt}
% \begin{wrapfigure}[11]{R}[0pt]{0pt}
% \centering
% %\raisebox{0pt}[\dimexpr\height-5\baselineskip\relax]{        \includegraphics[scale=0.4]{img/limitation.pdf}}%
%  \includegraphics[width=0.55\textwidth]{img/limitation.pdf}
% \caption{A failure case on a mirror-like object.}
% \label{fig:limitation}
% \end{wrapfigure}
% Although our method produces satisfactory results in most cases, it has several limitations.
% First, the projector pattern will not be captured by the cameras, and no correspondences can be obtained if the material of the object is mirror-like. In this case our method only relies on photometric supervision. In Fig.~\ref{fig:limitation} we show a failure case on a synthetic scene with a textureless and mirror-like reflection. Our method fails to reconstruct an accurate surface owing to the lack of structured-light supervision. It should be noted that this material is also challenging for other state-of-the-art methods.
% Second, although our method can optimize camera poses, it requires a reasonable camera pose initialization using markers or SfM softwares.

%% file: s5_conclusion.tex
\vspace{-1em}
\section{Conclusion}
\vspace{-0.5em}
% In the close-up of the dish, we can see that the SL method produced a double-layer surface. This occurred because the point clouds reconstructed from different camera pairs cannot be tightly aligned based on noisy camera poses. In addition, it may be observed that the shiny surface of the ash tray led to many error points around the surface for the SL method. By contrast, our proposed approach, benefiting from both DR- and SL-based methods, provided the most accurate 3D reconstruction results.

We proposed to supervise multi-view neural surface reconstruction by active sensing using structured light. Although existing neural reconstitution methods suffer from textureless surfaces, point clouds and multi-view correspondences obtained by structured-light provide sparse but more accurate supervision in such cases. On the other hand, structured-light systems are unsuitable for reflective surfaces and occlusions. These weaknesses are alleviated by the dense photometric supervision based on differentiable rendering. In experiments conducted on both synthetic and real-world datasets, we demonstrated that this combination significantly improves the performance of reconstructing challenging objects, and our method outperforms state-of-the-art neural surface reconstruction methods and conventional structured-light-based methods. We also found that the end-to-end self-calibration of camera poses enabled by our proposed loss functions is crucial for a high-quality reconstruction.